\documentclass[runningheads]{llncs}
\usepackage{graphicx}
\usepackage{xcolor}
\usepackage{hyperref}
\begin{document}
%
\title{Occupancy Estimation from Thermal Images}
%

%

\author{Zishan Qin\inst{1}
\and
Dipankar Chaki\inst{2}
\and
Abdallah Lakhdari\inst{2}
\and
Amani Abusafia\inst{2}
\and
Athman Bouguettaya\inst{2}
}
\authorrunning{Z. Qin et al.}
%
\institute{
School of Computing, Australian National University, Canberra, Australia \\
\email{taylor.qin2@anu.edu.au}\\
\and
School of Computer Science, University of Sydney, Australia\\
\email{\{dipankar.chaki,abdallah.lakhdari,amani.abusafia,athman.bouguettaya\}\\@sydney.edu.au}}

\maketitle              
\begin{abstract}

We propose a non-intrusive, and privacy-preserving occupancy estimation system for smart environments. The proposed scheme uses thermal images to detect the number of people in a given area. The occupancy estimation model is designed using the concepts of intensity-based and motion-based human segmentation. The notion of difference catcher, connected component labeling, noise filter, and memory propagation are utilized to estimate the occupancy number. We use a real dataset to demonstrate the effectiveness of the proposed system. 

\end{abstract}


\section{Introduction}

The emergence of intelligent technologies enables \emph{smart services} in the home environment to provide the residents with \emph{convenience} and \emph{efficiency} in our daily life \cite{naser2020adaptive}. 
Many research are based on a \emph{single occupant} environment \cite{du2019novel}. In reality, \emph{multiple occupants} live in a dwelling. Therefore, a new functional model is needed to determine the number of people in a space, referred to as \emph{occupancy estimation}. In application, it can help with heating, ventilation, and air conditioning systems control and even security monitoring in the smart buildings \cite{naser2020adaptive}.



This demo focuses on occupancy estimation using thermal images. Thermal imaging cameras are chosen because they offer advantages over intrusive sensors (such as RGB cameras) in terms of privacy protection and better than non-intrusive sensors (such as passive infrared (PIR) sensors) in terms of estimation results \cite{beltran2013thermosense}.
There are many challenges in this task. One challenge is that many objects, such as the CPU of a computer, are similar in temperature to a human \cite{chidurala2021occupancy}. Another challenge is that indoor environments may have excessive illustrations and far too much thermal noise \cite{naser2020adaptive}.
We propose a computationally efficient, non-intrusive, and privacy-preserving occupancy estimation system for the smart environment. Our proposed model predicts the number of occupants with an average accuracy of 71.6\% in six experiments by combining intensity-based and motion-based human segmentation.

\begin{figure}[t]
\centering \includegraphics[width=\linewidth]{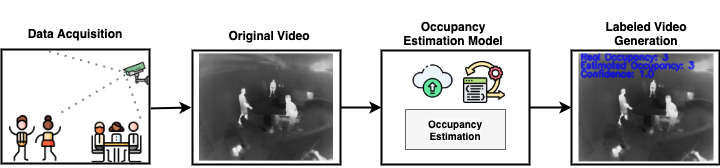}
\caption{The architecture of occupancy estimation system}
\label{system}
\end{figure}

\section{System Architecture}

The architecture of the system is shown in Fig. \ref{system}. The system has three major components: \emph{Data Acquisition}, \emph{Occupancy Estimation Model}, and \emph{Labeled Video Generation}. Thermal videos from an overhead view can be used as an input to our system. To guarantee the quality of the estimation results, the resolution of the video should be at least 200*100. Furthermore, people should not wear too thick clothes under the camera.
In the second component, human segmentation and classification are performed in the occupancy estimation model. Filters are used to determine the classes for the segments, and the number of selected classes determines the occupancy number. As output of our system, a labeled video is generated with predicted values indicating the number of people in the scene. From the video, each frame can be visually inspected by the user.


\section{Occupancy Estimation Model}

The proposed estimation model consists of two phases: \emph{Preliminary Parameter Configuration} and \emph{Occupancy Estimation}. A process flow that describes the process from the parameter configuration through the occupancy estimation is shown in Fig. \ref{framework}. We discuss each phase in the following subsections:

\begin{figure}
\centering\includegraphics[width=0.8\linewidth]{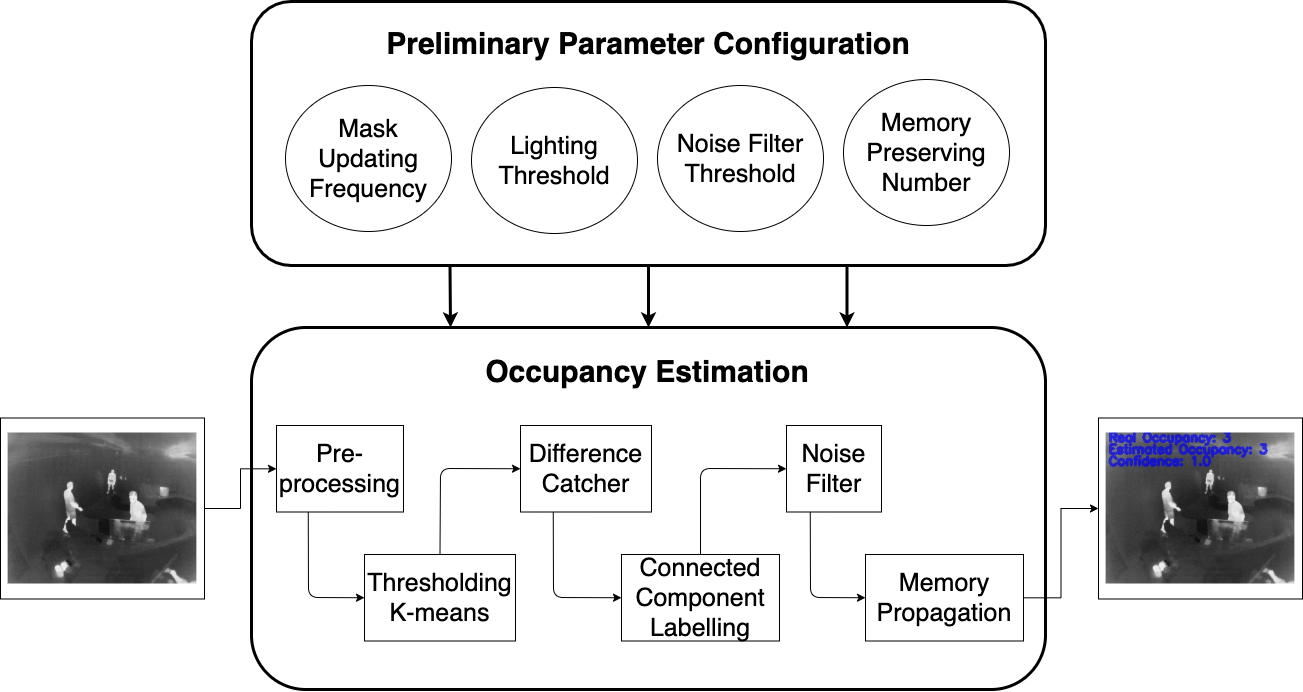}
\caption{Process flow diagram of the occupancy estimation model}
\label{framework}
\vspace{-4mm}
\end{figure}


\subsection{Preliminary Parameter Configuration}


In this phase, the main focus is to test the domain-specific parameters used in the estimation model via a binary search to get the suitable parameter combination. 
Several parameters include the mask updating frequency, lighting threshold, lower and upper bounds for the noise filter, and the memory preserving number. 

\vspace{-4mm}

\subsection{Occupancy Estimation}


This phase aims to use an efficient strategy to separate humans from other objects and then to calculate how many classes remain in the human layer through a classification algorithm. The main challenges include a large volume of thermal noises, either caused by the potential over-lighting or by other objects in a room with a similar temperature to humans. We perform the following steps to deal with these challenges for the occupancy estimation.

\vspace{-4mm}

\subsubsection{Pre-processing of the input video:} 
We divide the video into a sequence of images with an interval of two seconds for subsequent operations. Then, all pictures are cropped into a uniform resolution of 200*100 for faster runtime while maintaining important information. The first frame in the image sequence is stored into the model as an initialization mask, and this mask is updated according to the update frequency. The mask is used to reduce the effect of lighting variations between frames on the estimation results. 

\vspace{-4mm}


\subsubsection{Thresholding k-means:}
K-means allocates each point to the cluster with the nearest mean. This step first applies K-means to two consecutive frames to obtain two approximate segmentation results, respectively. Isolated thermal noises from over-segmentation are removed based on a lighting threshold from the preliminary configuration, followed by a Gaussian filter to blur the results. This reduces the side effects of hard thresholding. Then, we apply the mask from the previous step to eliminate the lighting issue in every two consecutive frames.

\vspace{-4mm}


\subsubsection{Difference catcher:} 
This step considers the motion between two consecutive frames. The difference is taken between two images generated from the prior step into a difference map. It is used to represent the dynamic change in the two consecutive frames which is caused by the movement of people.

\vspace{-4mm}


\subsubsection{Connected component labeling:} 
Connected component labeling is a commonly used method to detect connected regions in binary images \cite{he2017connected}. We apply this on the difference graph we get from the last step. Thus, we get a simulation of the dynamic differences in the actual movement of people in the room.


\vspace{-4mm}


\subsubsection{Noise filter:}
The output from the connected component may contain many small connected components, which are likely to be caused by thermal noises or over-illumination. That's why we need this step to eliminate the thermal noises. We use the thresholds from the parameter configuration phase to eliminate these small parts. Usually, we would not have a person occupying a large part of the overhead vision frame. In general, the low threshold is set relatively high when there are more other objects in the scene with similar temperatures to the human body. The more significant the lighting changes between these two images in the scene, the higher the high threshold will be.

\vspace{-4mm}


\subsubsection{Memory propagation:} 
This step prevents some outrageous predictions due to external factors. For example, the proposed model may over-segment when a person wears thicker clothes in the picture. The more complicated the indoor environment is, the smaller this memory propagation number will be.

\vspace{-2mm}



\section{Demo Setup}
\vspace{-1mm}
We use the Flir FB-Series thermal camera in our experimental setup, which has a relatively high resolution. The experiments have been conducted on a Mac operating system with a Core i3 processor and an 8GB RAM. Six experiments under different scenarios are done to test the effectiveness of our system, shown in Table \ref{tab:exp}. The output is a video of all frames of size 200*400, marked with the corresponding predicted occupancy, actual occupancy, and confidence scores calculated by $confidence_i = 1 - \left(\frac{estimation_i - real_i}{real_i}\right)$. We estimate the occupancy every two seconds, and the predicted occupancy number is shown in the display's top left corner. A video demonstrating the system can be found in the following link: \url{https://youtu.be/Av9BkB_ZZJc}. 

\begin{table}[!t]
  \centering
  \caption {Experimental Result}
\label {tab:exp}
\begin{tabular}{ |p{0.19\textwidth}|p{0.1\textwidth}|p{0.1\textwidth}|p{0.1\textwidth}|p{0.1\textwidth}|p{0.1\textwidth}|p{0.1\textwidth}|p{0.12\textwidth}| } 
 \hline
 Experiment & 1 & 2 & 3 & 4 & 5 & 6 & Average \\ 
 \hline
 Accuracy (\%) & 66.7 & 71.4 & 66.7 & 74.3 & 80.0 & 70.6 & 71.6 \\ 
  \hline
 Confidence & 0.833 & 1.258 & 0.918 & 0.863 & 1.000 & 1.118 & 0.998 \\
   \hline   
Environment & OL & OL & MP & LL, TN & OL & TN & \\
 \hline
     \end{tabular}
     \begin{tabular}{c}{
            \textbf{OL}: Over Lighting
            \textbf{MP}: Multiple People
            \textbf{LL}: Local Lighting
            \textbf{TN}: Thermal Noises
        }
     \end{tabular}
\end{table}




\section*{Acknowledgement}
\vspace{-1mm}
This research was partly made possible by DP160103595 and LE180100158 grants from the Australian Research Council. The statements made herein are solely the responsibility of the authors.

\vspace{-2mm}



%
%
\bibliographystyle{splncs04}
\bibliography{OccEst
.bib}

\begin{thebibliography}{1}
\providecommand{\url}[1]{\texttt{#1}}
\providecommand{\urlprefix}{URL }
\providecommand{\doi}[1]{https://doi.org/#1}

\bibitem{beltran2013thermosense}
Beltran, A., Erickson, V.L., Cerpa, A.E.: Thermosense: Occupancy thermal based
  sensing for hvac control. In: 5th ACM Workshop on ESEEB. pp.~1--8 (2013)

\bibitem{chidurala2021occupancy}
Chidurala, V., Li, X.: Occupancy estimation using thermal imaging sensors and
  machine learning algorithms. IEEE Sensors Journal  \textbf{21}(6),
  8627--8638 (2021)

\bibitem{du2019novel}
Du, Y., Lim, Y., Tan, Y.: A novel human activity recognition and prediction in
  smart home based on interaction. Sensors  \textbf{19}(20), ~4474 (2019)

\bibitem{he2017connected}
He, L., Ren, X., Gao, Q., Zhao, X., et~al.: The connected-component labeling
  problem: A review of state-of-the-art algorithms. Pattern Recognition
  \textbf{70},  25--43 (2017)

\bibitem{naser2020adaptive}
Naser, A., et~al.: Adaptive thermal sensor array placement for human
  segmentation and occupancy estimation. IEEE Sensors Journal  \textbf{21}(2),
  1993--2002 (2020)

\end{thebibliography}

\end{document}